\documentclass[manuscript,screen,review=false,natbib=false]{acmart}

\AtBeginDocument{%
  \providecommand\BibTeX{{%
    \normalfont B\kern-0.5em{\scshape i\kern-0.25em b}\kern-0.8em\TeX}}}

\usepackage{hyperref}

\usepackage[numbers]{natbib}
\usepackage{caption}
\usepackage[utf8]{inputenc} 
\usepackage[T1]{fontenc}    
\usepackage{hyperref}       
\usepackage{url}            
\usepackage{booktabs}       
\usepackage{amsfonts}       
\usepackage{nicefrac}       
\usepackage{microtype}      
\usepackage{float}
\usepackage{graphicx}
\usepackage{amsfonts}
\usepackage{booktabs}
\usepackage{siunitx}
\usepackage{algorithm}
\usepackage{ragged2e}
\usepackage{multirow}
\usepackage{colortbl}
\usepackage{hhline}
\usepackage{amsmath}
\usepackage{mhchem}
\usepackage{stmaryrd}
\usepackage{algpseudocode}
\usepackage{svg}
\usepackage{caption,subcaption}
\usepackage{lipsum}
\usepackage{float}
\usepackage{enumitem}
\usepackage{accents}

\newenvironment{blockquote}{%
  \par%
  \leftskip=2em%
   \topsep=0em
  }

\acmDOI{XXXXXXX.XXXXXXX}

\begin{document}

\title{Batch Layer Normalization\\ 
    A new normalization layer for CNNs and RNNs}

\author{Amir Ziaee}
\orcid{0000-0002-0648-489X }
\email{amir.ziaee@tuwien.ac.at}
\affiliation{%
  \institution{TU Wien}
  \streetaddress{Karlsplatz 13}
  \country{Austria}
}

\author{Erion Çano}
\affiliation{%
  \institution{University of Vienna}
  \country{Austria}
}

\renewcommand{\shortauthors}{Ziaee, et al.}

\begin{abstract}
This study introduces a new normalization layer termed Batch Layer Normalization (BLN) to reduce the problem of internal covariate shift in deep neural network layers. As a combined version of batch and layer normalization, BLN adaptively puts appropriate weight on mini-batch and feature normalization based on the inverse size of mini-batches to normalize the input to a layer during the learning process. It also performs the exact computation with a minor change at inference times, using either mini-batch statistics or population statistics. The decision process to either use statistics of mini-batch or population gives BLN the ability to play a comprehensive role in the hyper-parameter optimization process of models. The key advantage of BLN is the support of the theoretical analysis of being independent of the input data, and its statistical configuration heavily depends on the task performed, the amount of training data, and the size of batches. Test results indicate the application potential of BLN and its faster convergence than batch normalization and layer normalization in both Convolutional and Recurrent Neural Networks. The code of the experiments is publicly available
online.\footnote{\url{https://github.com/A2Amir/Batch-Layer-Normalization}}
\end{abstract}

\ccsdesc[500]{Computing methodologies~Neural networks}

\keywords{Neural Networks, Normalization, Internal Covariate Shift}

\maketitle

\section{Introduction}
Deep Neural Networks (DNNs) are extensively utilized across a wide range of domains, including Natural Language Processing (NLP), Computer Vision (CV), and Robotics applications. They usually consist of deep-stacked layers between which there is a linear mapping with learnable parameters and a non-linear activation function. Although their deep and complex structure gives them high representational capacity in learning features, it also makes them challenging to train where there is a randomness in the parameter initialization of layers and input data. This problem is called internal covariate shift \cite{ioffe2015batch} and occurs in the training of networks when previous layers' parameters change; the distribution of inputs for the current layer changes accordingly so that the current layer must constantly adapt to the new distributions. This problem is particularly severe for deep networks where small changes in deeper hidden layers are amplified as they propagate through the network, causing significant shifts in deeper hidden layers.
Many normalization methods have been introduced that have some disadvantages and advantages besides reducing the internal covariate shift. The disadvantages are accounting for a significant part of training time and playing no role in a hyper-parameter optimization process. On the other hand, advantages come from using a higher learning rate without vanishing or exploding gradients, acting as a regularizer so that the network enhances its generalization properties, speeding up training, and creating more reliable models.

In this study, we first provide a survey of the commonly used normalization methods, batch and layer normalization, in order to fully understand the advantages and disadvantages of each method. Next, we combine batch and layer normalization into one method termed Batch Layer Normalization (BLN), by which the drawbacks of each method are overcome, and a new normalization technique is developed that embodies the advantages of both methods. Unlike previous works that consider the role of normalization methods in the hyperparameter optimization process of a model as a relatively unexplored parameter, we additionally unify our method with the process of setting hyperparameters to fine-tune a model. This way, BLN as a normalization layer can play a comprehensive role in the hyperparameter optimization process of a model, leading to the stabilization and further improvement of the model.

Batch Layer Normalization, as a normalization layer during the learning process, estimates normalization statistics from the summed inputs to the neurons within a hidden layer across mini-batch and features. At inference times, it also performs the exact computation with a minor change, using either mini-batch statistics or population statistics. The decision to use either statistics of mini-batch or population is based on a model's hyper-parameter optimization process, giving the new normalization method the ability to play a role in the hyper-parameter optimization process. We show that BLN works well for the tasks that use convolutional or recurrent layers and improves the training time and the generalization performance of models.

\section{Related Work}
\label{sec:relatedwork}
Normalization methods can be classified into three categories \cite{gitman2017comparison}. The first category applies normalization to different dimensions of the output. Three popular instances for that are Layer Normalization \cite{ba2016layer}, in which inputs are normalized across features; Instance Normalization \cite{ulyanov2016instance}, which normalizes over the spatial locations of the output; and Group Normalization \cite{wu2018group}, that performs normalization independently along spatial dimensions and a group of features. The second category performs direct changes to the original batch normalization method \cite{ioffe2015batch}. This category includes methods such as Ghost BN \cite{hoffer2017train}, which performs normalization independently across different splits of batches, and Batch Re-normalization \cite{ioffe2017batch}, or Streaming Normalization \cite{liao2016streaming}, both of which make some changes to the original algorithm to use global averaged statistics rather than the current batch statistics. The final category includes methods based on normalizing weights instead of activations. This category consists of Weight Normalization \cite{salimans2016weight} and Normalization Propagation \cite{arpit2016normalization}. They all rely on dividing weights by their l2 norm and vary only in minor details. 

\begin{figure}[!h]
\includegraphics[width=13.5cm]{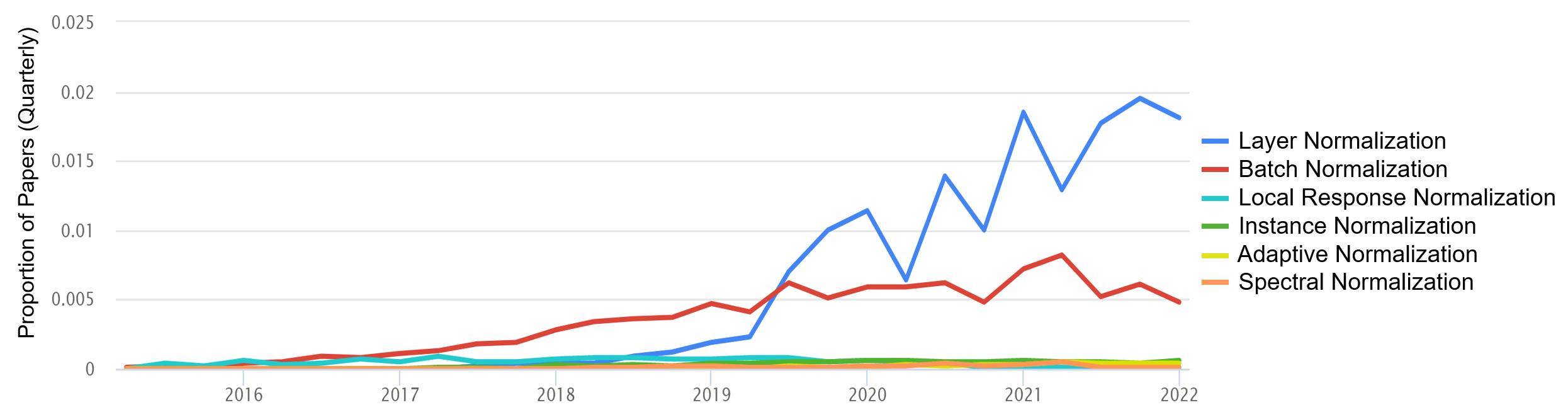}
\caption{Trend in using normalization methods \cite{PapersWithCode}}
\label{fig:image1}
\vspace*{-2mm}
\end{figure}

Despite the different normalization methods used in various fields, there is a trend in normalization methods employed by different papers over time, as shown in Figure \ref{fig:image1}.  The graph shows that two of the most commonly used normalization techniques are layer and batch normalization methods from the first and second categories. Accordingly, they became the fundamental component in most modern architectures \cite{szegedy2015going,zagoruyko2016wide,he2016identity,qi2017pointnet,huang2017densely,xie2017aggregated,szegedy2016rethinking} and transformers \cite{vaswani2017attention,yu2018qanet,xu2019understanding,xiong2020layer} and have made a successful spread in various areas of deep learning \cite{russakovsky2015imagenet,lin2014microsoft,chang2015shapenet}. 

Batch normalization can be used to manipulate the statistical properties of layer activations. Well-designed statistical properties can represent the domain-specific information for the distribution of a group of inputs. More explicitly, batch normalization tries to force the activations of a layer into a gaussian unit distribution at the beginning of training. It standardizes the activations of DNN intermediate layers with an approach of a three-step computation:\\
\textbf{1. Computation of Statistics:} the mean and variance of each mini-batch during the training phase are calculated. Let $x$ denote inputs over a minibatch $B$ of size $m, B=\left\{x_{1}, x_{2}, \ldots, x_{m}\right\}$, each input with a dimension of $\left(x^{(1)}, \ldots, x^{(d)}\right)$. Batch normalization \cite{ioffe2015batch} standardizes the mini-batch data by:
\begin{equation}
\mu_{{B}}^{({k})}=\frac{1}{{~m}} \sum_{{i}=1}^{{m}} {x}_{{i}}^{({k})}  \hspace{1cm}   \sigma_{{B}}^{({k})^{2}}=\frac{1}{{~m}} \sum_{{i}=1}^{{m}}\left({x}_{{i}}^{({k})} -\mu_{{B}}^{({k})}\right)^{2}
\end{equation}
Then, each dimension of the input is normalized separately:
\begin{equation}
\hat{{x}}_{{i}}^{({k})}=\frac{{x}_{{i}}^{({k})}-\mu_{{B}}^{({k})}}{\sqrt{\sigma_{{B}}^{({k})^{2}}+\epsilon}}
\end{equation}\\
where ${k} \in[1, {~d}]$ and ${i} \in[1, {~m}];$ $\mu_{{B}}^{({k})}$ and $\sigma_{{B}}^{({k})^{2}}$ are the per-dimension mean and variance, and $\epsilon>0$ is a small number to prevent numerical instability, respectively.\\
\textbf{2. Batch Normalization Transform}: batch normalization utilizes two additional learnable parameters to reconstitute a possible reduced representational capacity \cite{liao2016streaming}. Scale parameter $\gamma \in$ ${R}$ and shift parameter $\beta \in {R}$ where $\gamma$ and $\beta$ are subsequently learned in the optimization process.
\begin{equation}
{y}_{{i}}^{({k})}=\gamma^{({k})} \hat{{x}}_{{i}}^{({k})}+\beta^{({k})}
\end{equation}
The output of the transform step is then passed to other network layers, while the normalized output remains internal to the current layer.\\
\textbf{3. Inference:} the normalization steps depend on mini-batches in the training phase. In the inference phase, however, this dependence is no longer helpful. Instead, the normalization step in this phase is calculated deterministically. The population mean $E\left[x^{(k)}\right]$ and variance $\operatorname{Var}\left[x^{(k)}\right]$ are estimated by calculating the moving averages of the summed input statistics:
\begin{equation}
{E}\left[{x}^{({k})}\right]={E}_{{B}}\left[\mu_{{B}}^{({k})}\right]
\end{equation}
\begin{equation}
\operatorname{Var}\left[x^{(k)}\right]=\frac{m}{m-1} E_{B}\left[\sigma_{B}^{(k)^{2}}\right]
\end{equation}
Thus, the population statistics are a complete representation of all mini-batches. Therefore, the batch normalization transform in the inference step becomes:
\begin{equation}
{y}^{({k})}=\frac{\gamma^{({k})}}{\sqrt{\operatorname{Var}\left[{x}^{({k})}\right]+\epsilon}} {x}^{({k})}+\left(\beta^{({k})}-\frac{\gamma^{({k})} {E}\left[{x}^{({k})}\right]}{\sqrt{\operatorname{Var}\left[{x}^{({k})}\right]+\epsilon}}\right) \equiv {BN}_{\gamma^{(k)}, \beta^{(k)}}^{{inference}}\left({x}^{({k})}\right)
\end{equation}\\
where $y^{(k)}$ is passed to future layers instead of $x^{(k)}$. Since the parameters are fixed in this transformation, the batch normalization procedure applies a linear map to the activation.

Batch normalization has several advantages, including improving training stability, which is mainly due to its scale-invariant property \cite{gitman2017comparison,yong2020gradient,neyshabur2015data,wan2020spherical}, meaning a bad input from the previous layer does not ruin the next layer. Although the practical success of batch normalization is undeniable and is present in current state-of-the-art architectures pervasively, its theoretical analysis remains limited. For example, in the case of optimization, most analyses require the model to be independent of input data, such that the stochastic/mini-batch gradient becomes an unbiased estimator of the true gradient over a dataset. However, batch normalization typically does not fit this data-independent assumption, and its optimization usually depends on the sampling strategy and the mini-batch size \cite{russakovsky2015imagenet}. Furthermore, batch normalization still struggles with some problems in certain contexts. For example, the inconsistent operation of batch normalization between training and inference restricts its suitability in complex networks such as recurrent neural networks \cite{gitman2017comparison,cooijmans2016recurrent,laurent2016batch,kurach2019large,salimans2016improved,bhatt2019crossnorm}.

In order to address the drawbacks of batch normalization, layer normalization was introduced to estimate the normalization statistics directly from the summed inputs to the neurons within a hidden layer across all features. In layer normalization, all hidden units in a layer (mostly feature layer) have the same normalization statistics, which can be different in each training step \cite{gitman2017comparison}. In other words, batch normalization converts to layer normalization with only two steps:\\
\textbf{1. Computation of Statistics}: the mean and variance used for normalization are computed from all summed inputs to the neurons on each training step. Let $x$ denote inputs over a minibatch $B$ of size $m, B=\left\{x_{1}, x_{2}, \ldots, x_{m}\right\}$, each input with a dimension of $\left(x^{(1)}, \ldots, x^{(d)}\right)$, where ${i} \in[1, {~m}]$, and ${k} \in[1, {~d}]$. The mean and variance are computed as follows:
\begin{equation}
\mu_{F}^{(i)}=\frac{1}{d} \sum_{k=1}^{d} x_{i}^{({k})} \\ \hspace{0.95cm} \sigma_{F}^{({i})^{2}}=\frac{1}{d} \sum_{k=1}^{d}\left(x_{i}^{({k})} - \mu_{F}^{(i)}\right)^{2}
\end{equation}
Then, each sample is normalized such that the elements in the sample have zero mean and unit variance.  
\begin{equation}
\hat{x}_{{i}}^{({k})}=\frac{{x}_{{i}}^{({k})} -\mu_{F}^{(i)}}{\sqrt{\sigma_{F}^{({i})^{2}}+\epsilon}}
\end{equation}
\textbf{2. Transform}: by applying transformation on the normalized inputs using some learned parameters $\gamma$ and $\beta$, the output could be expressed as ${B}^{\prime}=\left\{{y}_{1}, {y}_{2}, \ldots, {y}_{{m}}\right\}$, where ${y}_{{i}}=$ ${LN}_{\gamma, \beta}\left({x}_{{i}}\right)$.
\begin{equation}
y_{i}=\gamma \hat{x}_{i}+\beta \equiv L N_{\gamma, \beta}\left(x_{i}\right)
\end{equation}

Layer normalization performs the exact computation at training and inference times and does not require the moving averages of summed input statistics. In contrast to batch normalization, layer normalization is not subject to any restriction regarding the size of mini-batches and can be used in pure online mode with the batch size of one. Layer normalization has also proven to be an effective method for stabilizing the hidden state dynamics in recurrent neural networks. From an empirical point of view, it shows that it can significantly reduce the training time compared to previously published techniques \cite{gitman2017comparison}. Despite the increasing usage of layer normalization in common neural network architectures \cite{vaswani2017attention,al2019character,yang1906generalized} employed for NLP, layer normalization works not as well as batch normalization when used with convolutional layers. When layers are fully connected, all layers' hidden units generally contribute similarly to the final prediction, and normalizing of summed inputs in a layer works well. However, similar contributions no longer hold for Convolutional Neural Networks \cite{gitman2017comparison}.

The following section describes the theoretical assumptions and practical experiments of developing the new normalization method in overcoming the drawbacks of batch and layer normalization methods and having a new normalization technique that embodies the advantages of both methods in the area of Convolutional Neural Networks (CNNs) and Recurrent Neural Networks (RNNs).

\section{Methodology}
This study introduces a new normalization mechanism called BLN, in which batch and layer normalization methods are combined into one method that overcomes the disadvantages of each method and embraces the advantages of both of them. In batch settings, where a model is trained on the steps of an entire training set, we would use the whole training set to normalize the activations, which is impractical in stochastic optimization. Since we use mini-batches in stochastic gradient training and each mini-batch has estimates for the mean and variance of each activation. Therefore, the batch normalization authors simplified and used mini-batches statistics for normalization. One issue they did not consider was the size of mini-batches. As discussed, batch normalization has the problem of small batch size, and its error increases rapidly as the batch size gets smaller \cite{ulyanov2016instance}. In contrast to batch normalization, layer normalization is not subject to any restriction regarding the size of mini-batches and can be used with the batch size of one.

\begin{figure}[!h]
  \vspace{-3mm}
  \centering{\includegraphics[width=13.5cm]{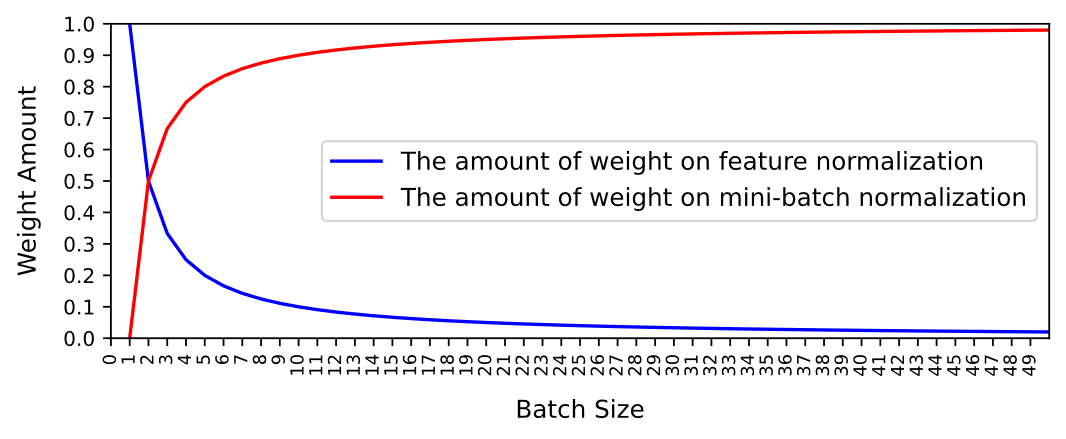}}
  \vspace*{-3mm}
  \caption{Effect of the numerator of Equation 16. on mini-batch and feature normalization.}
  \label{fig:image2}
  \vspace{-2mm}
\end{figure}

To address this limitation in our new normalization method, we consider the size of mini-batches in calculating normalization through one essential step, in which activations are independently normalized on mini-batch (Equation. 14) and features (Equation. 15) by having the mean equal to zero and the variance equal to one. Next, these normalized activations are combined based on a function (the numerator of Equation. 16) of the inverse size of mini-batches, which acts as a parameter in the range of $\epsilon$ to $1-\epsilon$ to put the appropriate weights on the normalized activations of mini-batch $\left(\hat{x}\right)$ and features $\left(\hat{\vphantom{\rule{1pt}{5.5pt}}\smash{\hat{x}}}\right)$. The function's behavior is shown in Figure \ref{fig:image2} for the size of mini-batches ranging from 1 to 50. As shown in Figure \ref{fig:image2}, increasing the size of the mini-batches increases the weight put on mini-batch normalization, while its decrease causes an increase in the amount of weight on feature normalization.

To adaptively control the scaling weights for different layers, we divided the numerator of Equation. 16 by the root
mean square of $d$ (the last dimensional shape of input), based on the idea presented in the AdaNorm algorithm \cite{xu2019understanding}. The issue of the normalizing equation (Equation. 16) is that a gradient descent optimization does not take into account the fact that normalizations occur. To solve this problem, the gradient of loss with respect to model parameters has to consider the normalizations and their dependence on model parameters. Therefore, we used a pair of learnable parameters $\gamma, \beta$ (Equation. 17) that scale and shift the normalized values for each input. Formally, let $x$ denote inputs over a minibatch $B$ of size $m, B=\left\{x_{1}, x_{2}, \ldots, x_{m}\right\}$, each input with a dimension of $\left(x^{(1)}, \ldots, x^{(d)}\right)$, where ${i} \in[1, {~m}]$, ${k} \in[1, {~d}]$, and $\epsilon>0$ is a small number to avoid numerical instability, $\gamma$ and $\beta$ are parameters to be learned. We compute output $\left\{y_{i}=B L N_{\gamma, \beta}\left(x_{i}\right)\right\}$ with Algorithm \ref{algorithm:algorithm1}.
\begin{algorithm}[!h]
\justifying
\noindent
\textbf{Input:} Inputs over a mini-batch: $B=\left\{x_{1}, x_{2}, \ldots, x_{m}\right\}$, each with a dimension of $\left(x^{(1)}, \ldots, x^{(d)}\right)$; Parameters to be learned: ${ \gamma, \beta}$ \\
\textbf{Output:} $\left\{y_{i}=B L N_{\gamma, \beta}\left(x_{i}\right)\right\}$\vspace{2mm}\\ 
{$\mu_{{B}}^{({k})} \leftarrow \frac{1}{{~m}} \sum_{{i}=1}^{{m}} {x}_{{i}}^{({k})}$       \hspace{9.4cm}           mini batch mean (10)}  \vspace{1mm}   \\
{$\sigma_{{B}}^{({k})} \leftarrow \sqrt{\frac{1}{{~m}} \sum_{{i}=1}^{{m}}\left({x}_{{i}}^{({k})} -\mu_{{B}}^{({k})} \right)^{2}+\epsilon}$        \hspace{5.6cm}   mini-batch standard   deviation (11)}\vspace{1mm}  \\
{$\mu_{F}^{(i)} \leftarrow \frac{1}{d} \sum_{k=1}^{d} x_{i}^{({k})}$    \hspace{10.05cm}    feature mean (12)}  \vspace{1mm}  \\
{$\sigma_{F}^{({i})}  \leftarrow \sqrt{\frac{1}{d} \sum_{k=1}^{d}\left(x_{i}^{({k})} - \mu_{F}^{(i)}\right)^{2}}$     \hspace{6.9cm}   feature standard deviation (13)} \vspace{1mm}   \\
{$\hat{x}^{{(k)}}_{{i}} \leftarrow  \frac{{x}_{{i}}^{({k})}-\mu_{{B}}^{({k})}}{\sigma_{{B}}^{({k})}}$    \hspace{9.35cm}  normalize mini-batch  (14)} \vspace{1mm}   \\
{$\hat{\vphantom{\rule{1.0pt}{5.5pt}}\smash{\hat{x}}}^{{(k)}}_{{i}} \leftarrow  \frac{{x}_{{i}}^{({k})} -\mu_{F}^{(i)}}{\sigma_{F}^{({i})}}$       \hspace{9.7cm}    normalize features  (15)} \vspace{1mm} \\
{$\hat{\vphantom{\rule{1pt}{5.5pt}}\smash{\hat{\hat{x}}}}_{i} \leftarrow \frac{ \left(\left(1-\left(\frac{1}{m}+\epsilon\right)\right) \hspace{0.05cm}   \hat{x}_{{i}}\right) + \left(\left(\frac{1}{m}-\epsilon\right)  \hspace{0.05cm} \hat{\hat{x}}_{{i}}\right)}{\sqrt{d}}$    \hspace{8.8cm}     normalize  (16)}\vspace{1mm}  \\
{${y}_{{i}} \leftarrow \gamma  \hspace{0.2cm} x\hat{\vphantom{\rule{1pt}{5.5pt}}\smash{\hat{\hat{}}}}_{i} +\beta \equiv {BLN} \gamma, \beta\left({x}_{{i}}\right)$    \hspace{8.8cm}   scale and shift (17)} \vspace{1mm}  
\caption{Batch Layer Normalization Transform.}
\label{algorithm:algorithm1}
\end{algorithm}

The authors of the batch normalization method generally assume that the normalization of activations depends on mini-batch statistics during training but is never essential during inference \cite{ioffe2015batch}. They used during inference, population statistics rather than mini-batch statistics. Various studies \cite{gitman2017comparison,cooijmans2016recurrent,laurent2016batch,kurach2019large,salimans2016improved,bhatt2019crossnorm} have been conducted, showing that using population statistics works well with CNNs, but they do not work well for NLP tasks that use RNNs \cite{cooijmans2016recurrent,shen2020rethinking} on account of the recurrent connection to previous time stamps. For example, in a study \cite{cooijmans2016recurrent} used moving batch statistics for different time steps of RNNs, it was found that the performance of a batch normalization method is significantly lower than that of a layer normalization method. We also had the experience that a change such as using batch statistics instead of a global average of summed input statistics in the inference phase of a batch normalization method affects the performance of a model. 

On the other hand, the authors of the layer normalization method assume that the normalization of activations depends only on mini-batch statistics during training and inference \cite{ba2016layer}. Thus, it can be inferred that the assumption that normalizing activations during inference must always depend on the population of the entire training set is not entirely correct. To address this issue, we also note that previous works have thoroughly investigated the role of normalization methods in models' hyper-parameter optimization processes as a relatively unexplored parameter. Therefore, we give Batch Layer Normalization a role in the hyper-parameter optimization process of a model by selecting between two assumptions for each of all four statistics (4$^{2}$ possible configurations) in Algorithm \ref{algorithm:algorithm1}. In the first assumption, for the normalization of activations during inference, mean or standard deviation is obtained from the corresponding estimated population statistics by setting True in one of the Equations $18, 19, 20, 21$ (see Appendix), where the expectations are over training mini-batches of size $m$ and $\mu_{B}, \sigma_{B}, \mu_{F}, \sigma_{F}$ are their sample means and standard deviations on mini-batches and features (Equations 10, 11, 12, 13). In the second, mean or standard deviation is only taken from the statistics of mini-batch during inference by selecting False in one of the Equations $18, 19, 20, 21$ (see Appendix). The selection process can be assumed as a hyper-parameter optimization process to fine-tune a model, leading to stabilization and further improvement of the model. Next, the normalization (Equations 22, 23, 24 in Appendix) is applied to each input, which is further composed with the scaling by $\gamma$ and shift by $\beta$ to form a single linear transform that replaces BLN($x_{i}$). The described procedure for training a Batch Layer Normalization network is summarized in Algorithm \ref{algorithm:algorithm2}.

As shown in Equation. 26, batch normalization \cite{ioffe2015batch} suggests that the normalization layer can be placed before activation functions such as sigmoid or ReLU for both fully-connected and convolutional layers. Therefore, they add the batch normalization transform before the nonlinearity ${g}(.)$, where ${W}$ and ${b}$ are learned parameters of a model and ${u}$ is the layer inputs.
\begin{equation}\tag{26}
{z}={g}({BN}({Wu}+{b}))
\end{equation}

In a study that shows the impact of recent advances in CNN architectures and learning methods \cite{mishkin2017systematic}, the authors conduct comparative experiments on the ImageNet dataset \cite{russakovsky2015imagenet} to understand whether a batch normalization method should be placed before or after the nonlinearity. The experiment showed that batch normalization placement after the nonlinearity results in better accuracy. Therefore, we apply the BLN transform after the nonlinearity $g(.)$ in contrast to batch normalization.
\begin{equation}\tag{27}
{z}={BLN}({g}({Wu}+{b}))
\end{equation}

\section{Experiments}
This section aims to first evaluate BLN to the internal covariate shift problem of deep learning layers, second to compare it to the application of batch and layer normalization method, and finally, draw conclusions from the results about the applicability of the new method for better solving the problem of internal covariate shift.
As mentioned earlier, batch normalization works well with CNNs and has the problem of not working with small batch sizes and RNNs; in contrast, layer normalization works well with RNNs and small batch sizes and has the disadvantage of not working well with CNNs. To test the applicability of our method, both with Convolutional and Recurrent Neural Networks and different batch sizes, two different challenging experiments were designed. The first is in the image classification area using the CIFAR-10 \cite{krizhevsky2009cifar} dataset, and the second is in the domain of sentiment analysis on the IMDB movie review dataset. Furthermore, to ensure the validity of the results in each experiment, we use identical architectures, training hyper-parameters, optimizations, and grid-search algorithms but different normalization methods. For all experiments, we set $\epsilon$ in all Equation to be 0.0001 and train all networks from scratch using an Adam optimizer with batch sizes of 1 and 25.
\subsection{Classifying CIFAR-10 using CNN}
\label{cnn}
\begin{figure}[!h] 
  \centering{\includegraphics[width=13.5cm,height=4.5cm]{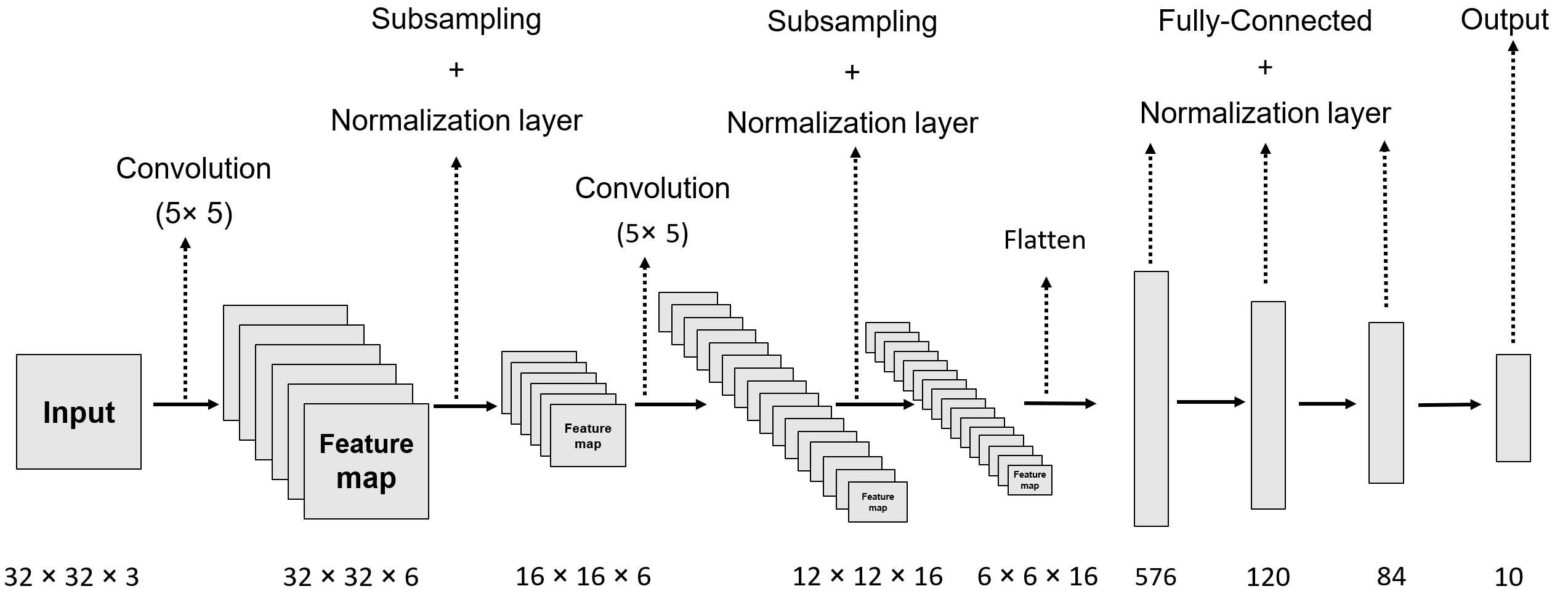}}
  \caption{Illustration of modified LeNet architecture.}
  \label{fig:cnnarchi}
  \vspace*{-2mm}
\end{figure}

To compare BLN against the other two approaches, batch and layer normalization in image classification and CNNs, we borrowed the popular LeNet-5 \cite{el2016cnn} architecture. Since there is no normalization layer in the original LeNet architecture, we derived a modified model from it (Figure \ref{fig:cnnarchi}) so that a normalization layer is embedded after some layers. By replacing the embedded normalization layers in the modified network with one of the discussed normalization methods, batch, layer, and BLN method, three independent networks were obtained in terms of normalization methods for each normalizer. Note that hyper-parameters and architectures used in training these networks are the same to ensure that the effect of these normalizers is compared appropriately.

\begin{figure}[!h]
\centering
\includegraphics[width=13.5cm, height=5.3cm]{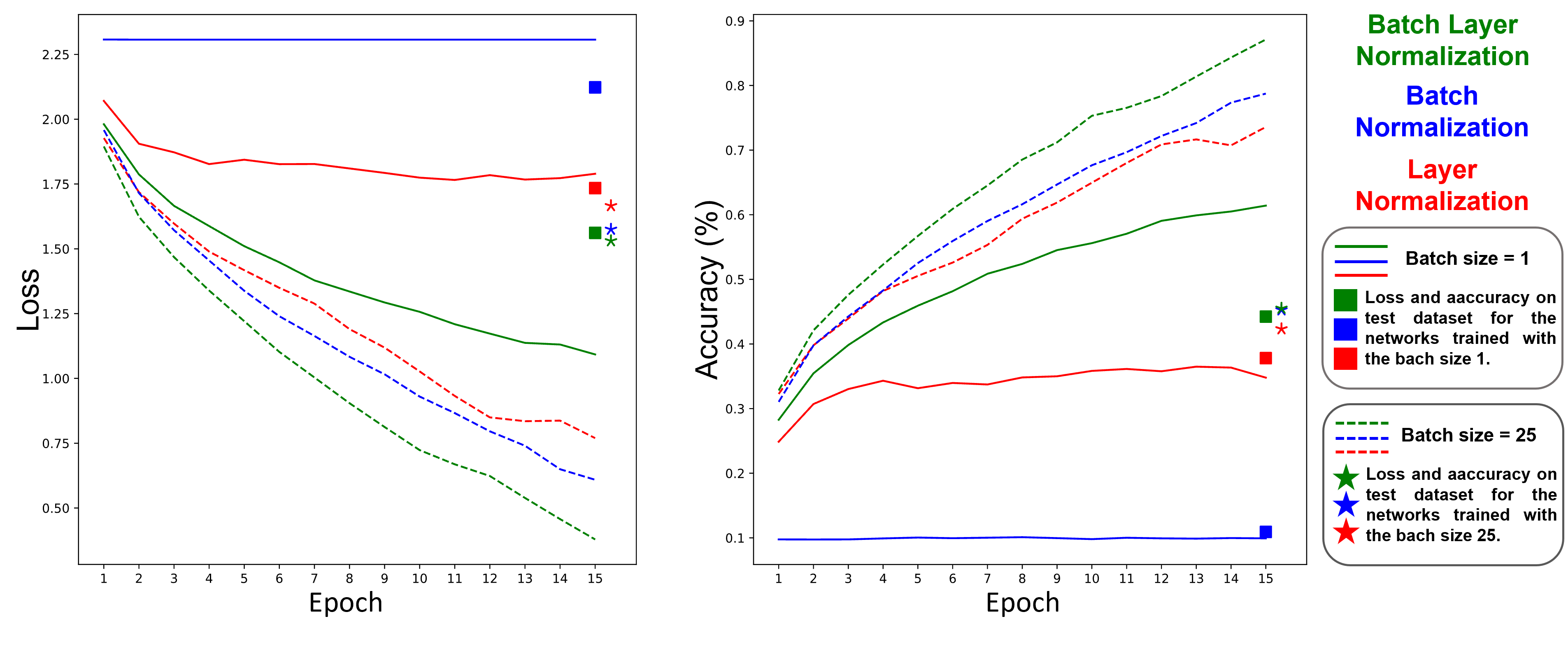} 
\vspace*{-3mm}
\begin{minipage}{6cm}
\centering
\vspace*{-4mm}
\subcaption{Train and Test loss}\label{fig:cnnres.a}
\end{minipage}%
\begin{minipage}{6cm}
\centering
\vspace*{-4mm}
\subcaption{Train and Test accuracy}\label{fig:cnnres.b}
\end{minipage}
\caption[]{Loss and Accuracy on training and test dataset.\label{fig:cnnres}}
\vspace{-2mm}
\end{figure}

As proof of the challenging experiments, after training each network independently on only \textbf{20\% of the training dataset} with batch sizes of 1 and 25, we compared their training results, including convergence issues and classification accuracy, as well as their performances on the \textbf{whole CIFAR-1O test set}. Loss results presented in Figure \ref{fig:cnnres.a} show that BLN significantly accelerates the neural network training and accomplishes a step toward reducing the internal covariant shift problem. In other words, considering the size of mini-batches in the calculation of normalization reduces the probability of getting stuck in the saturation regime and thus stimulates the training process. Figure \ref{fig:cnnres.a} also indicates that batch normalization with a batch size of 25 performs better than layer normalization. This finding notably confirms that in the case of big batch sizes, batch normalization has been the de facto standard for CNNs. Conversely, using batch normalization in CNNs significantly degrades the model's performance when batch size is one.

Based on Figure \ref{fig:cnnres.b}, which shows the experimental results in terms of the training and test accuracy, BLN enables the model to train faster and achieve higher accuracies of \textbf{0.61} and \textbf{0.87}, while batch normalization achieves 0.09 and 0.78 and layer normalization scores 0.34 and 0.73 on the training set with batch sizes 1 and 25, respectively. Moreover, BLN obtains better test accuracy compared to other normalizers.

Table \ref{table:table1} shows the possible configurations (step 12 of Algorithm \ref{algorithm:algorithm2}) for the BLN network trained on either the entire training dataset or 20\% of it with a batch size of 25. These configurations are found using the grid-search approach and sorted in terms of lower loss and higher accuracy on the entire test dataset. According to Table \ref{table:table1}, the training data size significantly impacts finding the best configuration of the BLN statistics in support of the theoretical analysis of independent input data and improving the practical results.

\begin{table}[!h]
\vspace{-1mm}
\centering
\caption{Configurations of BLN statistics used in CNN task.}
\vspace*{-3mm}
\label{table:table1}
\resizebox{12cm}{3.5cm}{%
\begin{tabular}{cccccccccc}
    \toprule

                             & \multicolumn{4}{c}{}                                                                                                                            &                              & \multicolumn{4}{c}{}                                                                                                                      \\
                             & \multicolumn{4}{c}{\multirow{-2}{*}{\begin{tabular}[c]{@{}c@{}}20\% of the training set used to train \\ with a batch size of 25\end{tabular}}} &                              & \multicolumn{4}{c}{\multirow{-2}{*}{\begin{tabular}[c]{@{}c@{}}Whole training set used to train\\  with a batch size of 25\end{tabular}}} \\
                                 \toprule

\multirow{-5}{*}{\textbf{A}} & \textbf{${E}_{B}$}                         & \textbf{$\operatorname{Std}_{B}$}                         & \textbf{${E}_{{F}}$}                        & \textbf{$\operatorname{Std}_{{F}}$}                        & \multirow{-5}{*}{\textbf{B}} & \textbf{${E}_{B}$}                       & \textbf{$\operatorname{Std}_{B}$}                       & \textbf{${E}_{{F}}$}                       & \textbf{$\operatorname{Std}_{{F}}$}                       \\
    \toprule

1                                            & {\cellcolor[rgb]{0,0.69,0.314}}True & {\cellcolor[rgb]{0,0.69,0.314}}True & {\cellcolor{red}}False              & {\cellcolor{red}}False              &                              & {\cellcolor{red}}False              & {\cellcolor[rgb]{0,0.69,0.314}}True & {\cellcolor{red}}False              & {\cellcolor{red}}False  \\
2                                            & {\cellcolor[rgb]{0,0.69,0.314}}True & {\cellcolor[rgb]{0,0.69,0.314}}True & {\cellcolor{red}}False              & {\cellcolor[rgb]{0,0.69,0.314}}True &                              & {\cellcolor{red}}False              & {\cellcolor[rgb]{0,0.69,0.314}}True & {\cellcolor{red}}False              & {\cellcolor[rgb]{0,0.69,0.314}}True  \\
3                                            & {\cellcolor[rgb]{0,0.69,0.314}}True & {\cellcolor[rgb]{0,0.69,0.314}}True & {\cellcolor[rgb]{0,0.69,0.314}}True & {\cellcolor[rgb]{0,0.69,0.314}}True &                              & {\cellcolor{red}}False              & {\cellcolor[rgb]{0,0.69,0.314}}True & {\cellcolor[rgb]{0,0.69,0.314}}True & {\cellcolor[rgb]{0,0.69,0.314}}True  \\
4                                            & {\cellcolor[rgb]{0,0.69,0.314}}True & {\cellcolor[rgb]{0,0.69,0.314}}True & {\cellcolor[rgb]{0,0.69,0.314}}True & {\cellcolor{red}}False              &                              & {\cellcolor{red}}False              & {\cellcolor[rgb]{0,0.69,0.314}}True & {\cellcolor[rgb]{0,0.69,0.314}}True & {\cellcolor{red}}False  \\
5                                            & {\cellcolor{red}}False              & {\cellcolor{red}}False              & {\cellcolor{red}}False              & {\cellcolor[rgb]{0,0.69,0.314}}True &                              & {\cellcolor{red}}False              & {\cellcolor{red}}False              & {\cellcolor{red}}False              & {\cellcolor{red}}False  \\
6                                            & {\cellcolor{red}}False              & {\cellcolor{red}}False              & {\cellcolor{red}}False              & {\cellcolor{red}}False              &                              & {\cellcolor{red}}False              & {\cellcolor{red}}False              & {\cellcolor{red}}False              & {\cellcolor[rgb]{0,0.69,0.314}}True  \\
7                                            & {\cellcolor{red}}False              & {\cellcolor{red}}False              & {\cellcolor[rgb]{0,0.69,0.314}}True & {\cellcolor[rgb]{0,0.69,0.314}}True &                              & {\cellcolor{red}}False              & {\cellcolor{red}}False              & {\cellcolor[rgb]{0,0.69,0.314}}True & {\cellcolor[rgb]{0,0.69,0.314}}True  \\
8                                            & {\cellcolor{red}}False              & {\cellcolor{red}}False              & {\cellcolor[rgb]{0,0.69,0.314}}True & {\cellcolor{red}}False              &                              & {\cellcolor{red}}False              & {\cellcolor{red}}False              & {\cellcolor[rgb]{0,0.69,0.314}}True & {\cellcolor{red}}False  \\
9                                            & {\cellcolor{red}}False              & {\cellcolor[rgb]{0,0.69,0.314}}True & {\cellcolor{red}}False              & {\cellcolor[rgb]{0,0.69,0.314}}True &                              & {\cellcolor[rgb]{0,0.69,0.314}}True & {\cellcolor[rgb]{0,0.69,0.314}}True & {\cellcolor{red}}False              & {\cellcolor[rgb]{0,0.69,0.314}}True  \\
10                                           & {\cellcolor{red}}False              & {\cellcolor[rgb]{0,0.69,0.314}}True & {\cellcolor{red}}False              & {\cellcolor{red}}False              &                              & {\cellcolor[rgb]{0,0.69,0.314}}True & {\cellcolor[rgb]{0,0.69,0.314}}True & {\cellcolor{red}}False              & {\cellcolor{red}}False  \\
11                                           & {\cellcolor{red}}False              & {\cellcolor[rgb]{0,0.69,0.314}}True & {\cellcolor[rgb]{0,0.69,0.314}}True & {\cellcolor[rgb]{0,0.69,0.314}}True &                              & {\cellcolor[rgb]{0,0.69,0.314}}True & {\cellcolor[rgb]{0,0.69,0.314}}True & {\cellcolor[rgb]{0,0.69,0.314}}True & {\cellcolor[rgb]{0,0.69,0.314}}True  \\
12                                           & {\cellcolor{red}}False              & {\cellcolor[rgb]{0,0.69,0.314}}True & {\cellcolor[rgb]{0,0.69,0.314}}True & {\cellcolor{red}}False              &                              & {\cellcolor[rgb]{0,0.69,0.314}}True & {\cellcolor[rgb]{0,0.69,0.314}}True & {\cellcolor[rgb]{0,0.69,0.314}}True & {\cellcolor{red}}False      \\
13                                           & {\cellcolor[rgb]{0,0.69,0.314}}True & {\cellcolor{red}}False              & {\cellcolor[rgb]{0,0.69,0.314}}True & {\cellcolor{red}}False              &                              & {\cellcolor[rgb]{0,0.69,0.314}}True & {\cellcolor{red}}False              & {\cellcolor[rgb]{0,0.69,0.314}}True & {\cellcolor{red}}False  \\
14                                           & {\cellcolor[rgb]{0,0.69,0.314}}True & {\cellcolor{red}}False              & {\cellcolor[rgb]{0,0.69,0.314}}True & {\cellcolor[rgb]{0,0.69,0.314}}True &                              & {\cellcolor[rgb]{0,0.69,0.314}}True & {\cellcolor{red}}False              & {\cellcolor[rgb]{0,0.69,0.314}}True & {\cellcolor[rgb]{0,0.69,0.314}}True  \\
15                                           & {\cellcolor[rgb]{0,0.69,0.314}}True & {\cellcolor{red}}False              & {\cellcolor{red}}False              & {\cellcolor[rgb]{0,0.69,0.314}}True &                              & {\cellcolor[rgb]{0,0.69,0.314}}True & {\cellcolor{red}}False              & {\cellcolor{red}}False              & {\cellcolor[rgb]{0,0.69,0.314}}True  \\
16                                           & {\cellcolor[rgb]{0,0.69,0.314}}True & {\cellcolor{red}}False              & {\cellcolor{red}}False              & {\cellcolor{red}}False              &                              & {\cellcolor[rgb]{0,0.69,0.314}}True & {\cellcolor{red}}False              & {\cellcolor{red}}False              & {\cellcolor{red}}False  \\   
    \toprule

\end{tabular}
}
\vspace{-1mm}
\end{table}

As shown in part B of Table \ref{table:table1}, the first-best possible configuration reveals that when using the whole training data, the current batch mean (${E}_{B}$ = False) and feature mean (${E}_{{F}}$ = False) are suitable settings to be used in the inference to normalize the input to layers of the network. The intuition is that the obtained global mean on batches and features may not share general information. In contrast, when using 20\% of the whole training set (part A of Table \ref{table:table1}), the first-best configuration suggests utilizing the population statistics on batches (${E}_{B}$ = True, $\operatorname{Std}_{B}$ = True) since we are using a subset of the training dataset and this information can assist the model in converging faster and learning better. Accordingly, this conclusion can be broadly applied to other statistic configurations of Table \ref{table:table1}. For example, the first-best and second-best configuration of the statistics show no advantage of using the population mean of features in both cases when the whole data or 20\% of it is used.

\subsection{Sentiment analysis of IMDB using RNN}
\label{rnn}
\begin{figure}[!h]
  \centering{\includegraphics[width=13.5cm, height=2.cm]{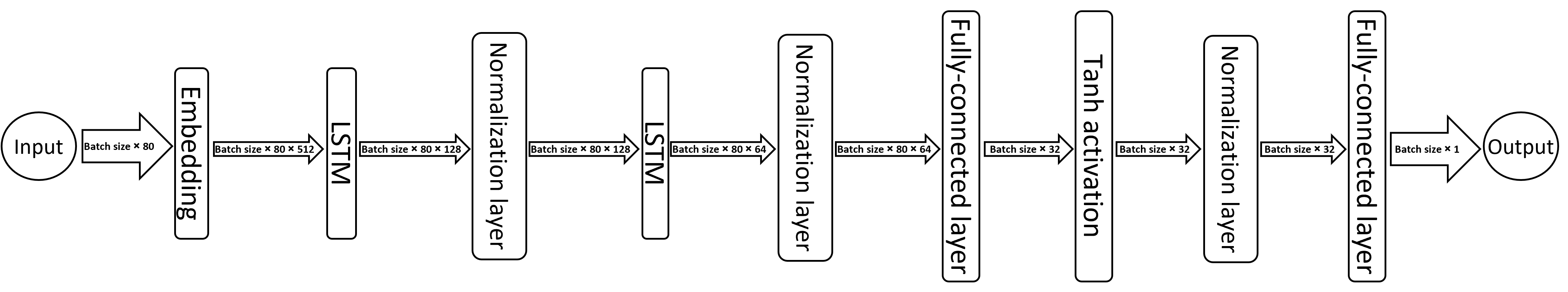}}
  \caption{Illustration of proposed RNN architecture.}
  \label{fig:fig5}
  \vspace*{-2mm}
\end{figure}

To investigate the performance and correlation between the statistical properties of BLN in the NLP domain, a sentiment analysis task was developed using an RNN with a focus on challenging the intriguing phenomenon that layer normalization is more effective than batch normalization in the NLP domain \cite{gitman2017comparison}, against the BLN method. Three independent networks were obtained by replacing all the normalization layers in the RNN architecture (Figure \ref{fig:fig5}) with one of the normalization methods discussed. These networks were trained and evaluated on the IMDB movie review dataset that contains 50,000 reviews divided equally across train and test splits. As a note, the experimental protocol described in Section \ref{cnn} was used for all experiments.

\begin{figure}[!h]
\centering
\includegraphics[width=13.5cm, height=5.3cm]{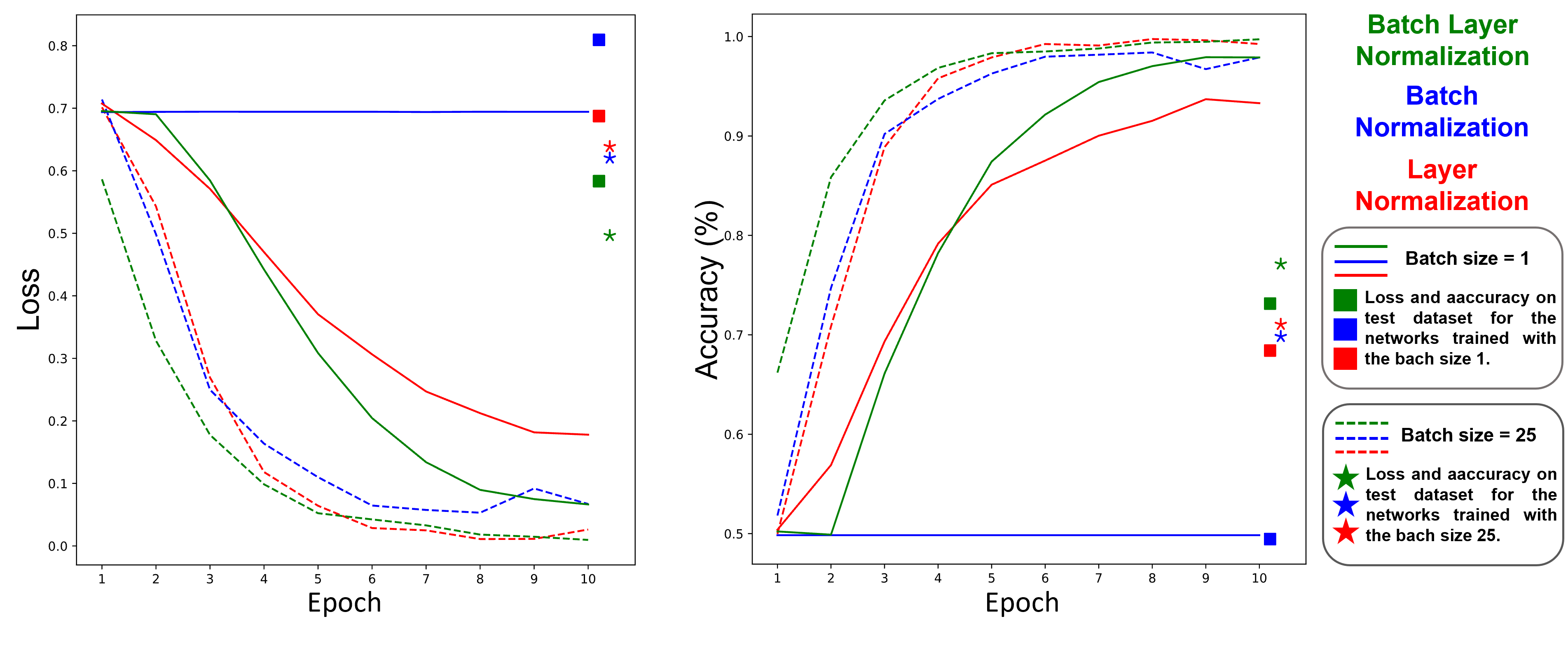} 
\vspace*{-3mm}
\begin{minipage}{6cm}
\vspace*{-4mm}
\subcaption{Train and Test loss}\label{fig:rnnres.a}
\end{minipage}%
\begin{minipage}{6cm}
\vspace*{-4mm}
\subcaption{Train and Test accuracy}\label{fig:rnnres.b}
\end{minipage}
\caption[]{Loss and Accuracy on training and test dataset.\label{fig:rnnres}}
\vspace*{-2mm}
\end{figure}

Figure \ref{fig:rnnres.a} and \ref{fig:rnnres.b} show the loss and accuracy of the obtained networks trained on \textbf{20\% of the training dataset} and evaluated on the \textbf{whole test dataset} with batch sizes of 1 and 25. As shown, BLN offers a per-iteration speedup and converges faster than batch normalization and even layer normalization, which is the most commonly used normalizer in the NLP domain. Based on the final results, our BLN normalizer shows significant improvement compared to the other two normalizers in all train and test evaluation metrics. 

As demonstrated in Figures \ref{fig:rnnres.a} and \ref{fig:rnnres.b}, the layer normalization potentially performs better than the batch normalization in terms of loss and accuracy. This result confirms the discussed assumption that layer normalization in RNNs improves the model's performance. Table \ref{table:table2}. summarizes the possible configurations (step 12 of Algorithm \ref{algorithm:algorithm2}) for the BLN network trained on either the entire training dataset or 20\% of it with a batch size of 25. These configurations are found using the grid-search approach and sorted in terms of lower loss and higher accuracy on the entire test dataset. Inferring from this table, we observe that the current batch mean (${E}_{{B}}$ = False) is a suitable configuration in the normalization process of the sentiment analysis task and elevates the performance while the global batch mean does not pass general information to the model or is not helpful at all for RNNs. In contrast, we recommend not using the global mean statistic obtained from the moving mean of batches when using either the whole training set or a subset of it. This deduction can broadly be extended to other sets of statistics.
\vspace*{-2mm}
\begin{table}[!h]
\vspace{-1mm}
\centering
\caption{Configurations of BLN statistics used in RNN task.}
\vspace*{-3mm}
\label{table:table2}
\resizebox{12cm}{3.5cm}{%
\begin{tabular}{cccccccccc}
 \toprule
                             & \multicolumn{4}{c}{}                                                                                                                            &                              & \multicolumn{4}{c}{}                                                                                                                      \\
                             & \multicolumn{4}{c}{\multirow{-2}{*}{\begin{tabular}[c]{@{}c@{}}20\% of the training set used to train \\ with a batch size of 25\end{tabular}}} &                              & \multicolumn{4}{c}{\multirow{-2}{*}{\begin{tabular}[c]{@{}c@{}}Whole training set used to train\\  with a batch size of 25\end{tabular}}} \\
                              \toprule
\multirow{-5}{*}{\textbf{A}} & \textbf{${E}_{B}$}                         & \textbf{$\operatorname{Std}_{B}$}                         & \textbf{${E}_{{F}}$}                        & \textbf{$\operatorname{Std}_{{F}}$}                        & \multirow{-5}{*}{\textbf{B}} & \textbf{${E}_{B}$}                       & \textbf{$\operatorname{Std}_{B}$}                       & \textbf{${E}_{{F}}$}                       & \textbf{$\operatorname{Std}_{{F}}$}                    \\
\toprule

1                                            & {\cellcolor{red}}False              & {\cellcolor{red}}False              & {\cellcolor[rgb]{0,0.69,0.314}}True & {\cellcolor{red}}False              &                                & {\cellcolor{red}}False              & {\cellcolor[rgb]{0,0.69,0.314}}True & {\cellcolor{red}}False              & {\cellcolor{red}}False               \\
2                                            & {\cellcolor{red}}False              & {\cellcolor{red}}False              & {\cellcolor{red}}False              & {\cellcolor{red}}False              &                                & {\cellcolor{red}}False              & {\cellcolor[rgb]{0,0.69,0.314}}True & {\cellcolor[rgb]{0,0.69,0.314}}True & {\cellcolor{red}}False               \\
3                                            & {\cellcolor{red}}False              & {\cellcolor{red}}False              & {\cellcolor[rgb]{0,0.69,0.314}}True & {\cellcolor[rgb]{0,0.69,0.314}}True &                                & {\cellcolor{red}}False              & {\cellcolor[rgb]{0,0.69,0.314}}True & {\cellcolor{red}}False              & {\cellcolor[rgb]{0,0.69,0.314}}True  \\
4                                            & {\cellcolor{red}}False              & {\cellcolor{red}}False              & {\cellcolor{red}}False              & {\cellcolor[rgb]{0,0.69,0.314}}True &                                & {\cellcolor{red}}False              & {\cellcolor[rgb]{0,0.69,0.314}}True & {\cellcolor[rgb]{0,0.69,0.314}}True & {\cellcolor[rgb]{0,0.69,0.314}}True  \\
5                                            & {\cellcolor[rgb]{0,0.69,0.314}}True & {\cellcolor{red}}False              & {\cellcolor{red}}False              & {\cellcolor[rgb]{0,0.69,0.314}}True &                                & {\cellcolor{red}}False              & {\cellcolor{red}}False              & {\cellcolor{red}}False              & {\cellcolor[rgb]{0,0.69,0.314}}True  \\
6                                            & {\cellcolor[rgb]{0,0.69,0.314}}True & {\cellcolor{red}}False              & {\cellcolor[rgb]{0,0.69,0.314}}True & {\cellcolor[rgb]{0,0.69,0.314}}True &                                & {\cellcolor{red}}False              & {\cellcolor{red}}False              & {\cellcolor[rgb]{0,0.69,0.314}}True & {\cellcolor[rgb]{0,0.69,0.314}}True  \\
7                                            & {\cellcolor[rgb]{0,0.69,0.314}}True & {\cellcolor{red}}False              & {\cellcolor{red}}False              & {\cellcolor{red}}False              &                                & {\cellcolor{red}}False              & {\cellcolor{red}}False              & {\cellcolor{red}}False              & {\cellcolor{red}}False               \\
8                                            & {\cellcolor[rgb]{0,0.69,0.314}}True & {\cellcolor{red}}False              & {\cellcolor[rgb]{0,0.69,0.314}}True & {\cellcolor{red}}False              &                                & {\cellcolor{red}}False              & {\cellcolor{red}}False              & {\cellcolor[rgb]{0,0.69,0.314}}True & {\cellcolor{red}}False               \\
9                                            & {\cellcolor{red}}False              & {\cellcolor[rgb]{0,0.69,0.314}}True & {\cellcolor[rgb]{0,0.69,0.314}}True & {\cellcolor{red}}False              &                                & {\cellcolor[rgb]{0,0.69,0.314}}True & {\cellcolor[rgb]{0,0.69,0.314}}True & {\cellcolor{red}}False              & {\cellcolor{red}}False               \\
10                                           & {\cellcolor{red}}False              & {\cellcolor[rgb]{0,0.69,0.314}}True & {\cellcolor{red}}False              & {\cellcolor{red}}False              &                                & {\cellcolor[rgb]{0,0.69,0.314}}True & {\cellcolor[rgb]{0,0.69,0.314}}True & {\cellcolor[rgb]{0,0.69,0.314}}True & {\cellcolor{red}}False               \\
11                                           & {\cellcolor{red}}False              & {\cellcolor[rgb]{0,0.69,0.314}}True & {\cellcolor[rgb]{0,0.69,0.314}}True & {\cellcolor[rgb]{0,0.69,0.314}}True &                                & {\cellcolor[rgb]{0,0.69,0.314}}True & {\cellcolor[rgb]{0,0.69,0.314}}True & {\cellcolor{red}}False              & {\cellcolor[rgb]{0,0.69,0.314}}True  \\
12                                           & {\cellcolor{red}}False              & {\cellcolor[rgb]{0,0.69,0.314}}True & {\cellcolor{red}}False              & {\cellcolor[rgb]{0,0.69,0.314}}True &                                & {\cellcolor[rgb]{0,0.69,0.314}}True & {\cellcolor[rgb]{0,0.69,0.314}}True & {\cellcolor[rgb]{0,0.69,0.314}}True & {\cellcolor[rgb]{0,0.69,0.314}}True  \\
13                                           & {\cellcolor[rgb]{0,0.69,0.314}}True & {\cellcolor[rgb]{0,0.69,0.314}}True & {\cellcolor{red}}False              & {\cellcolor{red}}False              &                                & {\cellcolor[rgb]{0,0.69,0.314}}True & {\cellcolor{red}}False              & {\cellcolor{red}}False              & {\cellcolor[rgb]{0,0.69,0.314}}True  \\
14                                           & {\cellcolor[rgb]{0,0.69,0.314}}True & {\cellcolor[rgb]{0,0.69,0.314}}True & {\cellcolor[rgb]{0,0.69,0.314}}True & {\cellcolor{red}}False              &                                & {\cellcolor[rgb]{0,0.69,0.314}}True & {\cellcolor{red}}False              & {\cellcolor[rgb]{0,0.69,0.314}}True & {\cellcolor[rgb]{0,0.69,0.314}}True  \\
15                                           & {\cellcolor[rgb]{0,0.69,0.314}}True & {\cellcolor[rgb]{0,0.69,0.314}}True & {\cellcolor{red}}False              & {\cellcolor[rgb]{0,0.69,0.314}}True &                                & {\cellcolor[rgb]{0,0.69,0.314}}True & {\cellcolor{red}}False              & {\cellcolor{red}}False              & {\cellcolor{red}}False               \\
16                                           & {\cellcolor[rgb]{0,0.69,0.314}}True & {\cellcolor[rgb]{0,0.69,0.314}}True & {\cellcolor[rgb]{0,0.69,0.314}}True & {\cellcolor[rgb]{0,0.69,0.314}}True &                                & {\cellcolor[rgb]{0,0.69,0.314}}True & {\cellcolor{red}}False              & {\cellcolor[rgb]{0,0.69,0.314}}True & {\cellcolor{red}}False      \\        
\toprule
\end{tabular}
}
\vspace{-1mm}
\end{table}

\section{Discussion}

The experimental results show that BLN converges significantly faster than batch and layer normalization in both CNN and RNN domains. Figures \ref{fig:cnnres} and \ref{fig:rnnres} also indicate that considering the size of mini-batches when calculating the normalization can contribute to finding the best statistical configuration, which leads to fast convergence of BLN. Furthermore, Tables \ref{table:table1} and \ref{table:table2} also show that the assumption that normalization of activations during inference must always depend on mini-batch statistics or population statistics is not a solid general assumption and must be modified according to the task, the amount of training data and the size of batches. These results notably approve that BLN supports the theoretical analysis of being independent of the input data by configuring its statistics used in the normalization process strongly based on the task, training data, and batch sizes. Complementary to this main point, other outcomes can be summarized as follows:

\begin{enumerate}
  \item From the first-best configuration in parts A and B of Table \ref{table:table1}, taking the global mean and  standard deviation of features (${E}_{{F}}$ = True, $\operatorname{Std}_{{F}}$ = True) in CNN normalization is not recommended regardless of training data size. In summary, the global batch mean is generally a good statistic for normalizing CNNs, when the training data is small, and vice versa; the current batch mean when the training data is big enough. 
  \item The first-best configurations in parts A and B of Table \ref{table:table2} declare that while utilizing the population mean on batches (${E}_{{B}}$ = True) in RNN normalization is not a good idea at all, the current mean and standard deviation of features can help RNNs converge faster.
  \item The first-best configurations in both parts (A and B) of Tables \ref{table:table1} and \ref{table:table2} approve that using the current standard deviation of features ($\operatorname{Std}_{{F}}$ = False) in normalization can usually help CNNs and RNNs converge faster.
\end{enumerate}

The limitation of BLN can be reflected in the time spent searching for the best configuration of statistics. This time can be reduced using the heuristic hyper-parameter tuning methods \cite{yu2020hyper}. In future work, step 12 of Algorithm \ref{algorithm:algorithm2}, which is tailored to overall neural network architecture, can also be extended to the statistics used for each deep neural network layer.

\section{Conclusions}

In this paper, we proposed a novel normalization method called BLN to reduce the internal covariate shift in deep learning layers and significantly accelerate their training. As a normalization layer, BLN not only exploits the advantages of the two most commonly used normalizers in the deep learning networks, batch, and layer normalization but also suppresses their drawbacks. Our proposed method derives its strength during training from appropriate weighting on mini-batch and feature normalization based on the inverse size of mini-batches and integrating this normalization in the network architecture. This guarantees that the normalization is appropriately handled by each optimization method used to train the network. During inferences times, it also performs the exact computation with a slight modification, using either mini-batch or population statistics as a decision process to play a comprehensive role in hyper-parameter optimization of models. BLN's main advantage is that it supports the theoretical analysis of being independent of the input data, as its statistical configuration is highly dependent on the task we are performing, the amount of training data, and the size of batches. We empirically verify that BLN is superior to using batch and layer normalization both in Convolutional and Recurrent Neural Networks, respectively, which further establishes the generalizability of our method in different deep learning models.

\newpage
\appendix
\section*{Appendix}
\setcounter{section}{1}
\begin{algorithm}[!h]
\justifying
\noindent
\textbf{Input}: Network $N$ with trainable parameters $\theta$; subset of activations $\left\{x^{(\mathbf{k})}_{i}\right\}_{i=1, k=1}^{m, d}$ \\
\textbf{Output}: Batch Layer Normalization network for inference, $N_{B L N}^{\inf }$ \vspace{1mm}\\
$1: {N}_{{BLN}}^{{tr}} \leftarrow {N}$   \hspace{7.8cm}  // Training BLN network   \\
$2:$ \textbf{for} $i=1 \ldots m$ do \\
\phantom{00000}$3:$ Add transformation $y_{i}=\operatorname{BLN}_{\gamma, \beta}\left(x_{i}\right)$ to $N_{B L N}{t {r}}($ Algorithm. \ref{algorithm:algorithm1})\\
\phantom{00000}$4:$ Modify each layer in $N_{B L N}^{\operatorname{tr}}$ with input $x_{i}$ to take $y_{i}$ instead\\
$5:$ \textbf{end for} \vspace{1mm}\\
$6:$ Train $N_{B L N}^{t r}$ to optimize the parameters $\Theta  \cup \left \{\gamma, \beta\right\}$\\
$7:$ $N_{B L N}^{\inf } \leftarrow N_{B L N}^{t r}$  \hspace{4cm}   // Inference BLN network with frozen parameters\\
$8:$ \textbf{for} $k=1 \ldots d$ do\\
\vspace{-3mm}
\begin{blockquote}$9:$ Set ${E}_{B}[{x}^{({k})}]=$ False, $\operatorname{Std}_{B}[{x}^{({k})}]=$ False, ${E}_{{F}}[{x}_{{i}}]=$ False, $\operatorname{Std}_{{F}}[{x}_{{i}}]=$ False \textbf{as initial parameters}. For clarity, \textbf{True} means using \textbf{population statistics} (process multiple training mini-batches ${B}$, each of size ${m}$ and average over them), \textbf{False} means using of \textbf{current mini-batch statistics}.
\end{blockquote}
$${E}_{B}[{x}^{({k})}]= \begin{cases}{E}\left[\mu_{B}^{({k})}\right] & \text { if True } \\ \frac{1}{{~m}} \sum_{{i}=1}^{{m}} {x}_{{i}}^{({k})} & \text {False}\end{cases}  \hspace{3.8cm} (18)
$$
$$\operatorname{Std}_{B}[{x}^{({k})}]= \begin{cases}\frac{{m}}{{m}-1} {E}\left[\sigma_{{B}}^{({k})}\right] & \text {if True} \\ \sqrt{\frac{1}{{~m}} \sum_{{i}=1}^{{m}}\left({x}_{{i}}^{({k})}-{E}_{{B}}[{x}^{({k})}]\right)^{2}+\epsilon} & \text {False}\end{cases}  \hspace{1.0cm} (19)
$$\\
\vspace{-3mm}
$${E}_{{F}}[{x}_{{i}}]= \begin{cases}{E}\left[\mu_{{F}}^{({i})}\right] & \text {if True} \\ \frac{1}{{~k}} \sum {x}_{{i}}^{({k})} & \text {False}\end{cases}  \hspace{4.7cm} (20)
$$\\
\vspace{-3mm}
$$\operatorname{Std}_{{F}}[{x}_{{i}}]= \begin{cases}\frac{{m}}{{m}-1} {E}\left[\sigma_{{F}}^{({i})}\right]  & \text {if True} \\ \sqrt{\frac{1}{k} \sum \left(x_{i}^{({k})} - {E}_{{F}}[{x}_{{i}}] \right)^{2}} & \text {False}\end{cases} \hspace{2.60cm} (21)
$$\\
$$\hat{x}^{{(k)}}_{{i}} \leftarrow \frac {{x}_{{i}}^{{k}}-{E}_{B}[{x}^{({k})}]}{\operatorname{Std}_{{B}}[{x}^{({k})}]} \hspace{5.7cm} (22)$$\\
\vspace{-3mm}
$$\hat{\hat{x}}^{{(k)}}_{{i}} \leftarrow \frac{{x}_{{i}}^{{k}} -{E}_{F}[{x}_{{i}}]}{\operatorname{Std}_{{F}}[{x}_{{i}}]}  \hspace{6.0cm} (23)$$\\
$10:$ \textbf{end for}\\
\phantom{}$11:$ In ${N}_{{BL} N}^{{inf}}$, replace the transform $y_{i}=\operatorname{BLN}_{\gamma, \beta}({x_{i}})$ with
$$ x\hat{\vphantom{\rule{1pt}{5.5pt}}\smash{\hat{\hat{}}}}_{i} \leftarrow \frac{ \left(\left(1-\left(\frac{1}{m}+\epsilon\right)\right) \hspace{0.05cm}   \hat{x}_{{i}}\right) + \left(\left(\frac{1}{m}-\epsilon\right)  \hspace{0.05cm} \hat{\hat{x}}_{{i}}\right)}{\sqrt{d}}    \hspace{3.45cm} (24)  \\$$ 
$${y}_{{i}} \leftarrow \gamma  \hspace{0.2cm} x\hat{\vphantom{\rule{1pt}{5.5pt}}\smash{\hat{\hat{}}}}_{i} + \beta   \hspace{6.95cm} (25) $$
$12:$ Use a grid-search algorithm or other hyper-parameter tuning techniques \cite{yu2020hyper} to find the best configuration of statistics (${E}_{B}$, $\operatorname{Std}_{B}$, ${E}_{{F}}$, $\operatorname{Std}_{{F}}$) among the possible configurations with lower loss and higher accuracy.\\
13: Use the best configuration for the rest of the training or fine-tuning and network testing.
\caption{Batch Layer Normalization Transform.}
\label{algorithm:algorithm2}
\end{algorithm}

\bibliographystyle{ACM-Reference-Format}
\bibliography{sample-base}

\end{document}